\pdfoutput=1
\documentclass[11pt]{article}

\usepackage[]{acl}
\usepackage{arydshln}
\usepackage{times}
\usepackage{latexsym}

\usepackage[T1]{fontenc}
\usepackage[utf8]{inputenc}

\usepackage{microtype}

\usepackage{inconsolata}

\usepackage{tikz}
\usepackage{graphicx}
\usepackage{multirow}
\usepackage{amsmath}
\usepackage{amssymb}
\usepackage{wrapfig}   % 用于实现文字环绕图片
\usepackage{float}
\usepackage{algorithm}
\usepackage{algorithmic}
\usepackage{mdframed}
\usepackage{tcolorbox}
\usepackage{subcaption}
\usepackage{listings}
\usepackage{lipsum}
\usepackage{array}
\usepackage{chngcntr}
\usepackage{ragged2e}
\tcbuselibrary{breakable}
\usepackage{booktabs}
\usepackage{makecell}
\usepackage{caption}
\usepackage{xcolor} % Added for color definitions
\usepackage{pifont}  % For \ding{55} (a cross)
\usepackage{longtable}
\usepackage{comment}
\usepackage{graphicx}
\usepackage{subcaption}
\usepackage[export]{adjustbox}

\usepackage{booktabs}
\usepackage{tabularx}
\usepackage{array}
\usepackage{makecell}
\usepackage{amssymb}

\usepackage{todonotes}

\usepackage[table]{xcolor} % 用于表格颜色     % 用于单元格内换行
\definecolor{lightgray}{gray}{0.9} % 定义一个浅灰色
\tcbuselibrary{skins}
\usepackage{diagbox}

\usepackage{xstring}

\author{
    Yijun Chen$^{1}$\thanks{ \ \ Equal contribution.}, 
    Boyi Xiao$^{1}$\footnotemark[1], Yixian Zhao$^{1}$\footnotemark[1], 
    Yi He $^{1}$\footnotemark[1], Xiangyu Guo$^{1}$\footnotemark[1], 
    Haoting Xia$^{1}$\footnotemark[1], \\
    \textbf{Buqiang Xu}$^{1}$, \textbf{Jizhan Fang}$^{1}$, 
    \textbf{Yanya Li}$^{1}$, \textbf{Yaqi Zheng}$^{1}$,
    \textbf{Xuehai Wang}$^{1}$, \\
    \textbf{Zirui Xue}$^{1}$, 
    \textbf{Liuxin Zhang}$^{2}$, \textbf{Hui Li}$^{2}$,
    \textbf{Ningyu Zhang}$^{1}$\thanks{ \ \ Corresponding author.} \\
    $^{1}$Zhejiang University \\
    $^{2}$Lenovo Group Limited \\
    \texttt{ie\_yijunchen@zju.edu.cn, zhangningyu@zju.edu.cn}
}

\definecolor{myGreen}{RGB}{0, 150, 0}
\definecolor{myRed}{RGB}{200, 0, 0}

\definecolor{myblue}{RGB}{30, 144, 255} % 纯正亮蓝，不发紫

\usepackage{bm}
\newcommand{\perf}[2]{%
    #1%
    \rlap{$
        \,_{\IfBeginWith{#2}{-}%
            {\color{myGreen}\text{\tiny{(#2)}}}%
            {\color{myRed}\text{\tiny{(#2)}}}%
        }
    $}%
}

\tcbuselibrary{most} 

\newtcolorbox{definitionbox}[1][]{%
  enhanced,                % 启用高级绘图引擎
  title={Definition},      % 默认标题
  colback=blue!5,          % 内容背景色
  colframe=blue!50!black,  % 边框颜色
  coltitle=white,          % 标题文字颜色
  colbacktitle=blue!60!black, % 标题背景色
  boxrule=1.5pt,           % 边框粗细
  arc=3mm,                 % 圆角大小 (统一设为 3mm 会比较好看)
  fonttitle=\bfseries,     % 标题字体
  
  attach boxed title to top left={yshift=-2mm,xshift=4mm}, 
  boxed title style={
    rounded corners,
    boxrule=0pt,           % 标题框不要边框，看起来更干净
  },
  
  breakable,               % 建议加上这个，防止长定义无法跨页
  #1                       % 允许调用时覆盖参数
}

\usepackage{utfsym}
\usepackage{stfloats}
\newtcolorbox{insightbox}{
    float*=!hb,            % 强制跨栏浮动到底部
    width=\textwidth,    % 宽度占满整页
    colback=myblue!5!white, % 极淡背景 (截图看起来像白底，你可以改成 white)
    fontupper=\small\linespread{1.05}\selectfont,
    colframe=black,         % 黑色细边框    
    boxrule=0.5pt,          % 边框很细 (0.5pt)
    arc=6pt,                % 圆角很小 (2pt)，看起来很锐利但有圆润感
    left=8pt, right=8pt, top=4pt, bottom=4pt,
    before skip=10pt,
    after skip=10pt
}

\usepackage[scale=0.85]{FiraMono}
\newtcolorbox{promptbox}[2][]{
  enhanced,
  title={#2},
  colframe=myblue!50!gray,
  colback=myblue!5!white,
  colbacktitle=myblue!50!gray,
  fonttitle=\bfseries\large,
  fontupper=\ttfamily\small\linespread{1.1}\selectfont, % 稍微减小行距更紧凑
  boxrule=1pt,
  arc=2mm,
  breakable,                % 允许跨页
  top=10pt, bottom=10pt,    % 增加一点内边距，防止文字贴边
  oversize,                 % 允许稍微突破栏宽（可选）
  #1
}

\tcbuselibrary{raster}
\usepackage{fontawesome5}

\newtcolorbox{CaseStudyFrame}[2]{
    enhanced, 
    title={\large \textbf{#1}},
    colback=white, colframe=gray!70!black, coltitle=white,
    fonttitle=\bfseries,
    width=\textwidth, 
    boxrule=1pt,
    arc=4pt,
    label={#2}
}

\newtcolorbox{SolutionColumn}[4]{
    enhanced,
    equal height group={#4}, % 外框等高
    title={\textbf{#3}},
    colframe=#1, colback=white, coltitle=white,
    subtitle style={colback=#2, colupper=black},
    drop shadow, 
    width=\linewidth,
    fonttitle=\small\bfseries,
    left=3pt, right=3pt,
    valign=top 
}

\newtcolorbox{ReportFrame}[2]{
    enhanced,
    title={\large \textbf{#1}}, % 标题
    colback=white, 
    colframe=gray!70!black, 
    coltitle=white,
    fonttitle=\bfseries,
    width=\textwidth,
    boxrule=1pt,
    arc=4pt,
    fontupper=\ttfamily\small\linespread{1.1}\selectfont, 
    label={#2}
}

\newtcolorbox{AlignedDesc}[1]{
    enhanced,
    frame hidden,      % 隐藏边框
    colback=white,     % 背景透明/白
    equal height group={#1_desc}, % 关键：使用后缀 _desc 自动分组
    left=0pt, right=0pt, top=0pt, bottom=5pt, % 调整间距
    nobeforeafter      % 防止不必要的换行干扰
}

\usepackage[scale=0.85]{FiraMono} % scale=0.85 防止字体过大

\newtcblisting{caserawbox}[2][]{
  enhanced,
  title={#2},
  colback=white,
  colframe=gray!70!black,
  coltitle=white,
  fonttitle=\bfseries\large,
  boxrule=1pt,
  arc=2mm,
  breakable,
  listing only,
  listing options={
    basicstyle=\ttfamily\scriptsize,
    breaklines=true,
    columns=fullflexible,
    keepspaces=true,
    showstringspaces=false,
    extendedchars=false, 
    literate={ }{ {~} }1 {≈}{{$\approx$}}1, 
    inputencoding=utf8,
  },
  #1
}

\usepackage{booktabs}
\usepackage{graphicx}
\usepackage{xcolor}
\usepackage{colortbl}

\definecolor{domainbg}{gray}{0.95}

\usepackage{xspace}

\title{
\texorpdfstring{%
\raisebox{-0.32\height}{\includegraphics[height=2.3em]{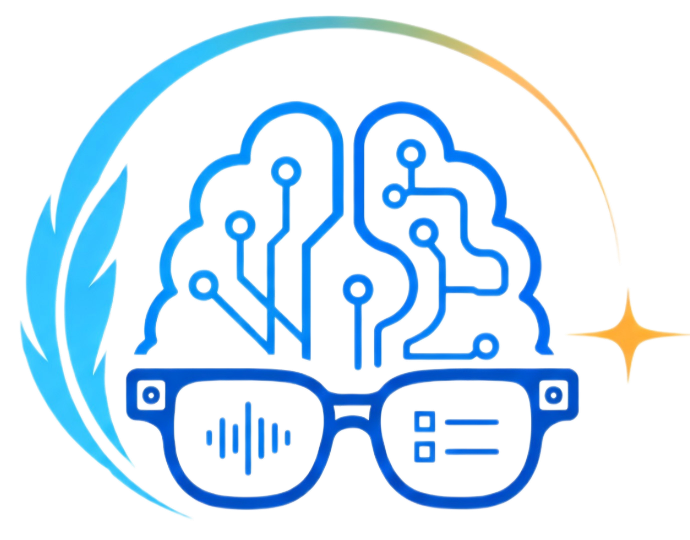}}%
\hspace{0.38em}%
LightMem-Ego: Your AI Memory for Everyday Life%
}{LightMem-Ego: Your AI Memory for Everyday Life}
}
\begin{document}
\maketitle

\begin{abstract}
Personal AI assistants on mobile and wearable devices continuously perceive users' daily lives through visual and audio streams. However, answering queries about past experiences requires lightweight multimodal memory that can continuously accumulate, organize, and retrieve long-term experiences, which remains challenging. To address this challenge, we present \textbf{LightMem-Ego}, a lightweight streaming multimodal memory system for everyday-life assistance. The system continuously captures egocentric visual and audio streams, aligns them on a shared timeline, and organizes them into a hierarchical memory consisting of current, short-term, and long-term memory. Given a user query, LightMem-Ego dynamically routes retrieval to the appropriate memory level and generates answers grounded in multimodal evidence. The demonstration can be deployed on smartphones and AI glasses, supporting object finding, conversation recall, life summarization, routine discovery, and personalized assistance\footnote{\url{https://github.com/zjunlp/LightMem-Ego}.}.
%LightMem-Ego demonstrates how lightweight long short-term multimodal memory enables personal AI assistants to move beyond one-shot perception toward continuous, experience-grounded interaction
\end{abstract}

\begin{comment}
sec1  Introduction

1 生活记忆的价值   眼镜  手机

2 挑战

3 我们做了啥

seC2  系统设计  

讲  里面哪些模块

sec3 应用场景   这里直接贴几个场景 按应用贴case

sec4 定量实验   自己构造一些case 和其他app 对比

sec5 总结展望

提出：

MemGlass。

支持跨分钟、天、周级别记忆问答。

2 Related Work
Conversational Memory

ChatGPT Memory

MemGPT

Mem0

Wearable AI

Meta Smart Glasses

Humane

Multimodal Memory

Episodic Memory

Experience Replay

Life Logging

3 MemGlass
3.1 Multimodal Life Capture

视频

音频

位置

屏幕

传感器

3.2 Event Segmentation

将连续生活流切分为：

Meeting
Walking
Dining
Shopping
Conversation
3.3 Hierarchical Memory

短期记忆

长期记忆

语义记忆

情景记忆

3.4 Experience Retrieval

时间感知检索

人物感知检索

地点感知检索

事件感知检索

3.5 Experience QA

从记忆生成答案。

4 Demonstration
Demo A

Lost Object Retrieval  我把钥匙放在哪里了？  1. 物品遗失与位置记忆

我钥匙放哪了？
我的耳机在哪里？
我昨天把电脑充电器放哪了？
我刚刚把手机放哪了？
我上次用完之后放回原位了吗？  

Demo B

Meeting Recall     2. 人际交互回忆（会议 / 对话 / 社交）

非常高频，尤其是工作场景。

典型问题：
开会记录 
我刚刚和谁说过这件事？
上周和老板讨论了什么？
这个人我见过吗？
我上次和他聊的项目是什么？
他是不是让我改过proposal？
我刚刚看到的那条信息是什么？
刚才那个网页讲了什么？
我刚刚刷到的视频在哪？
上次那个PPT讲的关键点是什么？

Weekly Life Summary  一周

5 Evaluation
Memory Retrieval Accuracy

Top-k Recall

MRR

Experience QA Accuracy

GPT-4o Judge

Human Evaluation

Latency

Mobile

Phone

Glasses

6 Conclusion

构建面向真实世界个人AI助手的长期多模态记忆系统，使AI从“理解语言”迈向“理解人生经历”。

\end{comment}

\begin{figure*}[t]
    \centering
    \includegraphics[width=\textwidth]{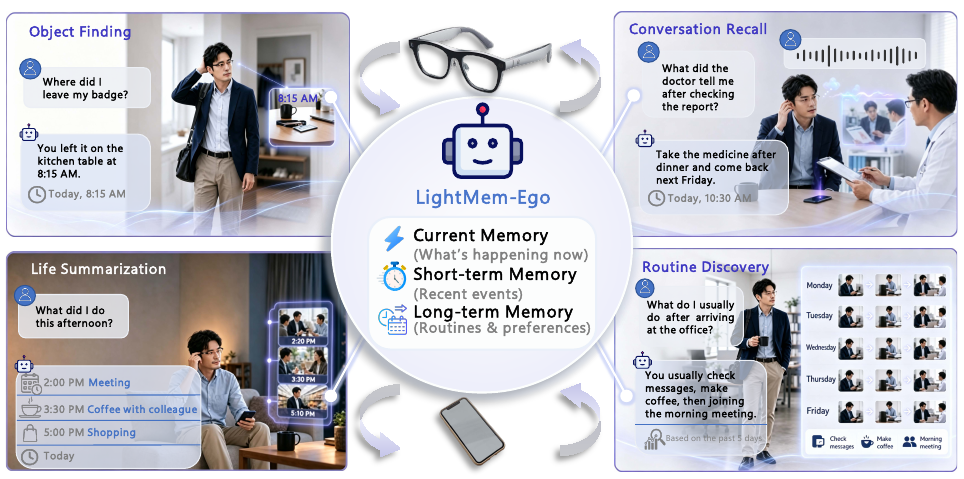}
    \caption{
    Overview of \textbf{LightMem-Ego}'s motivating scenarios and memory hierarchy. 
    The system supports everyday memory assistance across object finding, conversation recall, life summarization, and routine discovery by routing user queries to current, short-term, 
    and long-term memory.
    }
    \label{fig:intro_overview}
    \vspace{-5mm}
\end{figure*}

\section{Introduction}
Personal memory is a fundamental part of everyday intelligence. 
People constantly rely on memory to recall where they placed an object, what someone just said, what happened during a meeting, or how their routines change over time. 
As smartphones and AI glasses become natural interfaces for capturing egocentric visual and audio streams, they create a new opportunity for AI assistants: instead of only answering isolated questions, an assistant can become a memory companion for everyday life \cite{Vinci, EgoLife, Egocentric_Co-Pilot}. 
Such a system could help users find misplaced items, recall recent conversations, summarize daily activities, and reason over long-term habits and routines \cite{EgoLife, EgoMemory, LifeEval, MobileMem}.

Recent multimodal large language models have made human-computer interaction more natural by combining vision, speech, and language \cite{M3-Agent, WorldMM, GPT-4o, Gemini-2.5-pro}. 
Meanwhile, wearable devices and smartphones provide practical channels for continuous, hands-free perception in daily environments \cite{Vinci, EgoLife, Egocentric_Co-Pilot, SUPERGLASSES, MobileMem, deng2026mobilemem}.
However, turning such interfaces into everyday memory assistants requires addressing three fundamental challenges:
\textbf{First}, egocentric experience arrives as continuous visual-audio streams without explicit event boundaries, requiring assistants to transform raw observations into coherent event-level experiences.
\textbf{Second}, continuously accumulated experiences must be incrementally organized into current, short-term, episodic, and semantic memory while remaining efficient for long-term deployment.
\textbf{Third}, user queries naturally span multiple temporal horizons, requiring dynamic routing across different memory levels instead of relying on a single context window or flat retrieval store.
Together, these challenges call for an experience-centric memory system that can continuously capture, organize, retrieve, and reason over everyday life \cite{EgoMemory, MyEgo, EgoMemReason, SuperMemory-VQA, LifeEval}.

To address these challenges, we present \textbf{LightMem-Ego}, a deployable streaming multimodal memory system for everyday-life assistance. 
LightMem-Ego connects lightweight smartphone or AI-glasses-style clients with a backend that ingests egocentric visual-audio streams, segments recent observations into events, and organizes them into a three-level memory hierarchy: current memory for ongoing context, short-term memory for recent micro-events, and long-term memory for consolidated episodes and semantic facts \cite{MemGPT, Mem0, LightMem, StructMem, qiao2026delta}. 
During question answering, a memory router selects evidence according to the temporal scope and intent of the query, enabling grounded responses about the present, the recent past, and long-term routines within a unified interface. 
The system supports continuous frame streams, audio chunks, synchronized timestamps, and modular ASR, vision-language, language model, retrieval, and storage components. 
We demonstrate LightMem-Ego on representative egocentric memory tasks, including object finding, conversation recall, life summarization, and routine discovery (Figure~\ref{fig:intro_overview}), and report quantitative results on retrieval, question answering, and latency against application-oriented baselines without explicit multimodal long- and short-term memory.

\section{Related Work}

\paragraph{Conversational Memory.}
Conversational memory extends LLM-based assistants from stateless, single-session interaction to persistent agents that can retain, update, and retrieve user-specific information over time \cite{Memory_for_Autonomous_LLM_Agents, MemOS, LightMem, Mem0, wang2026learning}. 
Product systems such as ChatGPT Memory and research systems such as MemGPT and Mem0 demonstrate complementary designs: the former emphasizes personalized continuity, while the latter study explicit memory management through hierarchical context, external storage, consolidation, and retrieval~\cite{GPT-4o, GPT-5, Mem0, MemGPT, TokenPilot}. 
These systems motivate LightMem-Ego's use of persistent, queryable memory, but they mainly focus on language-centric interaction rather than streaming egocentric multimodal experience.

\paragraph{Wearable and Mobile AI Assistants.}
Wearable and mobile AI assistants place multimodal interaction in the user's physical environment. 
Wearable systems and smart-glasses agents use egocentric perception for real-time situated assistance \cite{Vinci, Vinci_new, EgoLife, Egocentric_Co-Pilot, Are_we_ready}. 
Related benchmarks and prototypes further evaluate glasses-style agents and connect visual understanding with physical-world actions \cite{SUPERGLASSES, VisualClaw, VisionClaw, ProMemAssist, EgoLife}. 
On smartphones, multimodal assistants increasingly combine language, vision, voice, and on-screen understanding \cite{GPT-4o, Gemini-2.5-pro}. 
However, most systems still focus on current-context perception and response, rather than explicit long-horizon memory of everyday experience.

\paragraph{Multimodal Personal Memory.}
Multimodal personal memory studies how daily experience can be captured, organized, and queried over time. 
Memory-augmented video agents build intermediate representations for temporally grounded reasoning over long videos \cite{VideoAgent, M3-Agent, WorldMM, Ego-R1}. 
Personalized egocentric retrieval and VQA recover user-specific events from long-context recordings \cite{EgoMemory, MyEgo, EgoMemReason, SuperMemory-VQA}. 
Lifelogging and egocentric-life systems organize daily archives for retrieval and assistance \cite{Lifelong_Retrieval, EgoLife}, while life-journaling systems move toward semantic activity summarization from mobile sensing and language models \cite{AutoLife, EM2Mem, EM²Mem}. 
These works motivate LightMem-Ego's streaming hierarchical memory design.

\begin{figure*}[t!]
    \centering
    \captionsetup[subfigure]{font=footnotesize,skip=2pt}

    % Row 1: Live Capture & Wearable Deployment (a)-(c)
    \begin{subfigure}[t]{0.548\textwidth}
        \centering
        \includegraphics[width=\linewidth]{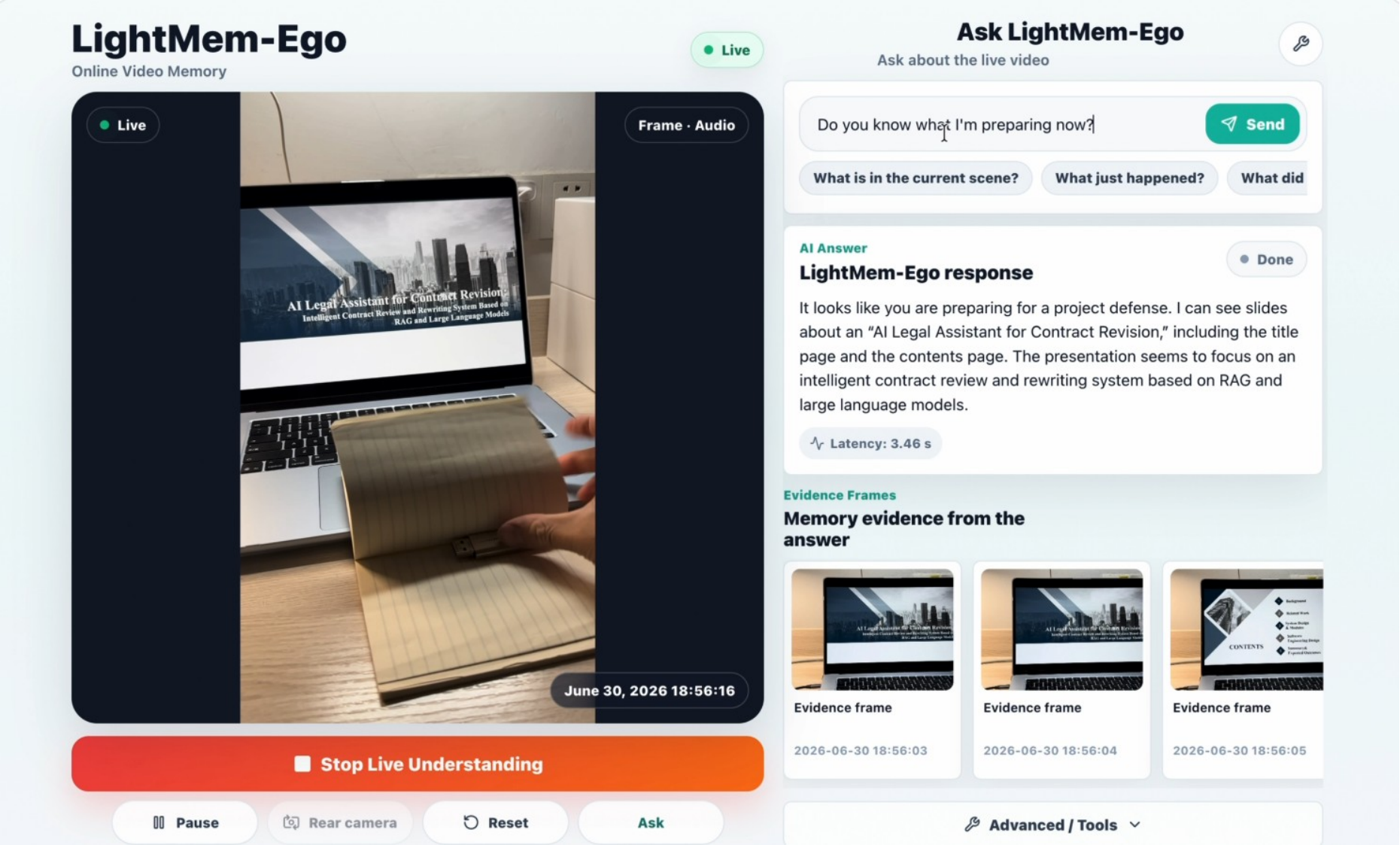}
        \caption{Web client (live capture \& QA).}
        \label{fig:system-interface-web}
    \end{subfigure}\hfill
    \begin{subfigure}[t]{0.234\textwidth}
        \centering
        \includegraphics[width=\linewidth]{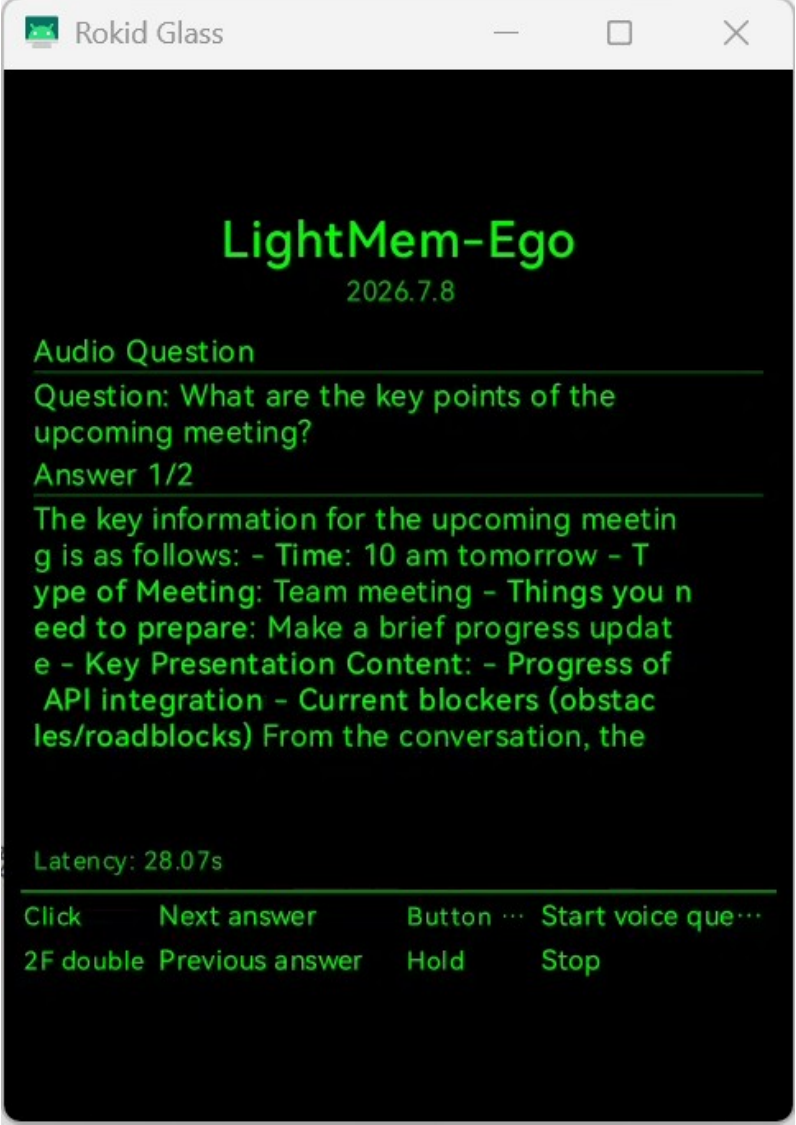}
        \caption{Glasses app UI.}
        \label{fig:system-interface-glass-app}
    \end{subfigure}\hfill
    \begin{subfigure}[t]{0.188\textwidth}
        \centering
        \includegraphics[width=\linewidth]{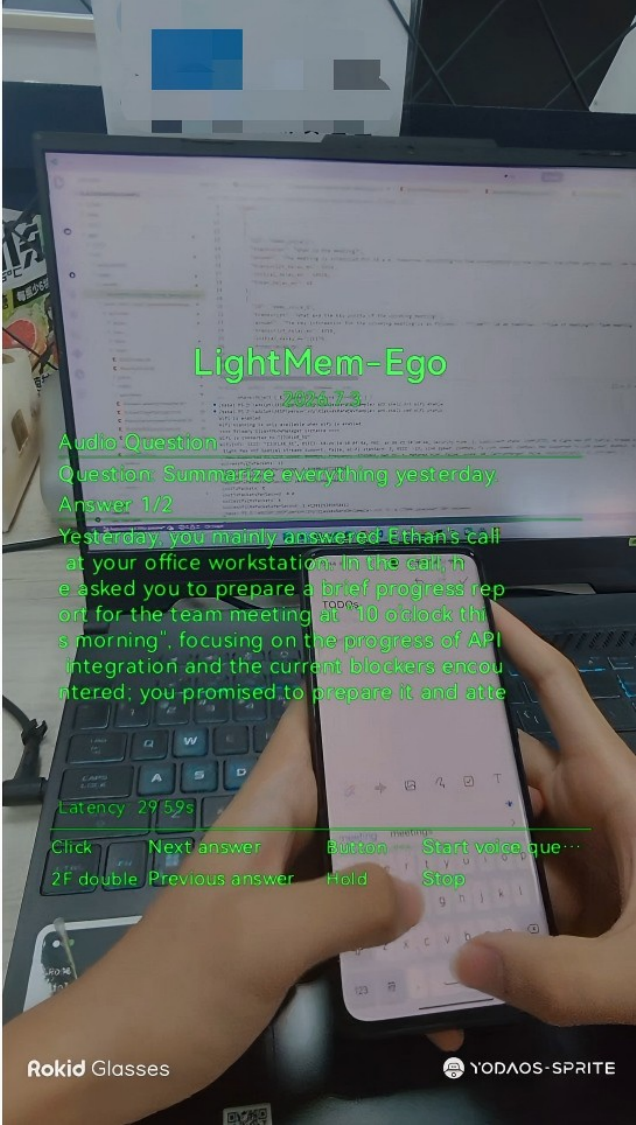}
        \caption{First-person overlay.}
        \label{fig:system-interface-glass-view}
    \end{subfigure}

    \vspace{0.55em}

    % Row 2: Memory Management Workspace (d)-(e)
    \begin{subfigure}[t]{0.541\textwidth}
        \centering
        \includegraphics[width=\linewidth]{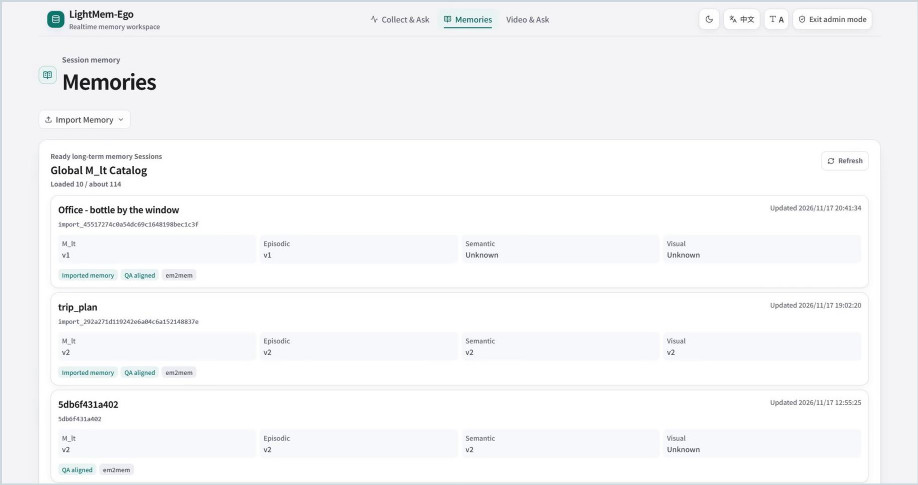}
        \caption{Memory catalog \& session import.}
        \label{fig:system-interface-mem-catalog}
    \end{subfigure}\hfill
    \begin{subfigure}[t]{0.444\textwidth}
        \centering
        \includegraphics[width=\linewidth]{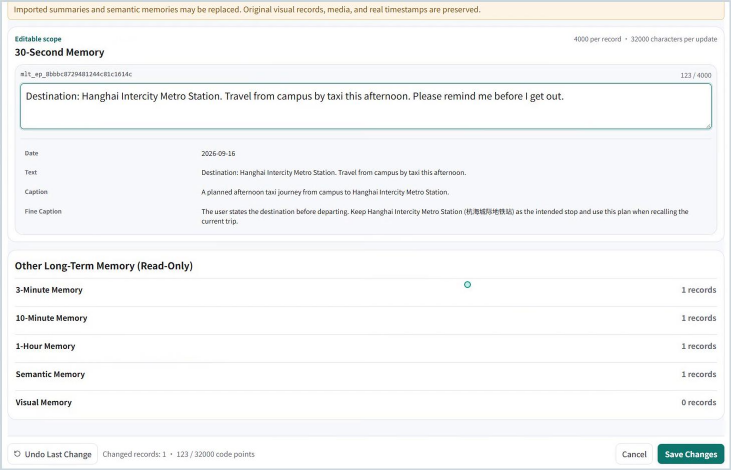}
        \caption{Editable long-term memory.}
        \label{fig:system-interface-mem-edit}
    \end{subfigure}

    \vspace{0.55em}

    % Row 3: Dual-Graph Visualization & Provenance (f)-(g)
    \begin{subfigure}[t]{0.487\textwidth}
        \centering
        \includegraphics[width=\linewidth]{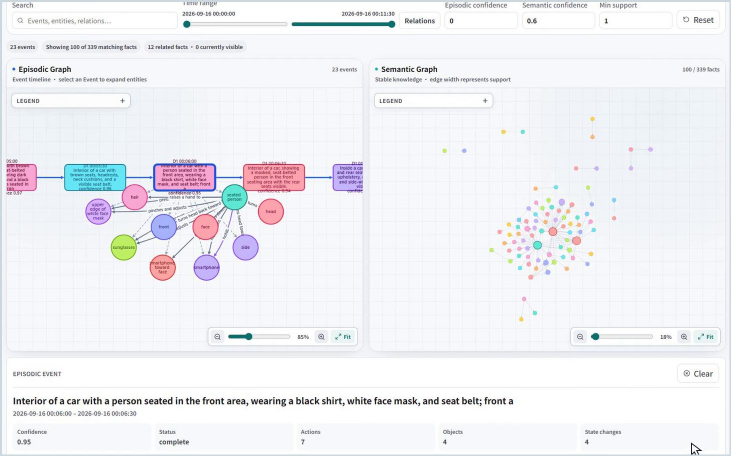}
        \caption{Episodic graph \& event inspection.}
        \label{fig:system-interface-mem-episodic}
    \end{subfigure}\hfill
    \begin{subfigure}[t]{0.498\textwidth}
        \centering
        \includegraphics[width=\linewidth]{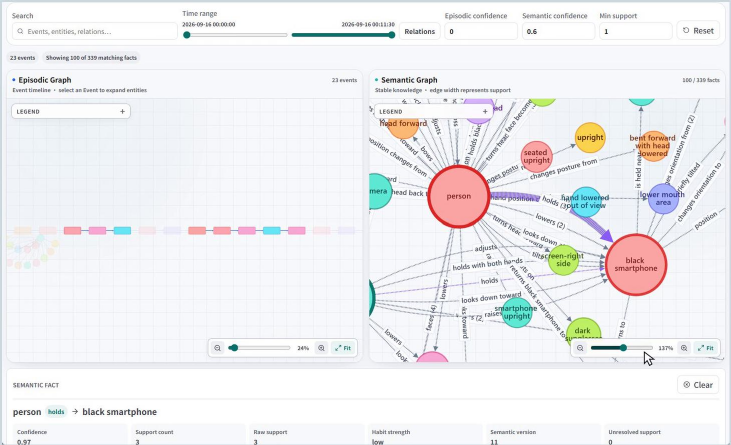}
        \caption{Semantic graph \& supporting facts.}
        \label{fig:system-interface-mem-semantic}
    \end{subfigure}

    \vspace{-0.3em}
    \caption{
    Interfaces of \textbf{LightMem-Ego} across live deployment and memory management.
    The top row (a--c) shows live multimodal capture, QA, and wearable egocentric responses;
    the middle and bottom rows (d--g) present the memory catalog, editable long-term records, and interactive episodic and semantic graph exploration.
    }
    \label{fig:system_interfaces}
    \vspace{-1.0em}
\end{figure*}

\section{LightMem-Ego}

LightMem-Ego is a streaming multimodal memory system for everyday-life assistance. 
It connects smartphones or wearable devices with a backend that captures multimodal life streams, organizes them into hierarchical memory, and answers user questions through explicit memory retrieval.
The system combines online updates for current and short-term memory with asynchronous consolidation into long-term episodic and semantic stores, allowing it to support both immediate scene understanding and retrospective reasoning over past experience.
Figure~\ref{fig:system_interfaces} shows the deployed interfaces of LightMem-Ego across live web, glasses-client, and first-person wearable views (a--c), together with the interactive memory management workspace for catalog browsing, memory editing, and episodic/semantic graph inspection (d--g).

\subsection{Multimodal Life Capture}

We represent everyday experience as a temporally ordered multimodal stream:
\begin{equation}
\mathcal{X}=\{x_t\}_{t=1}^{T}, \qquad x_t=(v_t,a_t,m_t),
\end{equation}
where \(v_t\), \(a_t\), and \(m_t\) denote visual observations, audio observations, and auxiliary metadata, respectively. 
The capture layer is lightweight: it collects, normalizes, and timestamps raw inputs from smartphones or wearable devices, and forwards them to the backend without heavy client-side reasoning. 
In online deployment, the primary interface is a unified frame--audio stream, in which image frames and short audio chunks are uploaded to the same session. 
All modalities are aligned on a shared session timeline using the relative timestamp \(\tau=t-t_0\), which is preserved for downstream event building, transcript backfilling, and retrieval. 
Beyond video and audio, the same interface can incorporate side-channel context such as coarse location, screen activity, or lightweight sensor cues.

\subsection{Event Segmentation}

Given the captured stream \(\mathcal{X}\), LightMem-Ego incrementally partitions it into short event segments 
\(\mathcal{E}=\{e_i\}_{i=1}^{N}\), where each segment corresponds to a contiguous interval on the shared timeline. 
Segmentation is driven by temporal continuity and cross-frame change signals, enabling lightweight micro-event construction under streaming conditions without semantic parsing of every frame. 
Each segment stores its temporal span, representative frames, provisional visual description, and aligned or pending audio context. 
As more evidence arrives, transcripts and refined descriptions are asynchronously attached to existing segments. 
These segments form the basic units for downstream memory construction and can represent everyday experiences such as meetings, walking episodes, shopping moments, or conversations.

\subsection{Hierarchical Memory}

LightMem-Ego organizes captured experience into a hierarchical memory system:
\begin{equation}
\mathcal{M}=\{\mathcal{M}_{cur},\mathcal{M}_{st},\mathcal{M}_{lt}\},
\end{equation}
where \(\mathcal{M}_{cur}\), \(\mathcal{M}_{st}\), and \(\mathcal{M}_{lt}\) denote current, short-term, and long-term memory, respectively. 
\(\mathcal{M}_{cur}\) acts as a lightweight working memory over the most recent multimodal observations and active event state, enabling immediate responses about the ongoing scene. 
\(\mathcal{M}_{st}\) stores recent event segments together with representative visual evidence, provisional or refined descriptions, and transcript context when available. 
Stable short-term events are then consolidated into long-term memory \(\mathcal{M}_{lt}\), which contains episodic memory \(\mathcal{M}_{epi}\) for event-centered past experiences and semantic memory \(\mathcal{M}_{sem}\) for higher-level regularities such as routines, preferences, and relationships. 
This hierarchy separates fast online interaction from slower memory consolidation while preserving both event-specific and abstract personal knowledge.

\subsection{Edge-Oriented Efficiency}

LightMem-Ego targets mobile and AI-glasses deployment by keeping edge processing lightweight and shifting memory construction to the backend.
The client samples, compresses, timestamps, and uploads low-rate frames and short audio chunks, avoiding heavy VLM or LLM inference on resource-constrained devices.
The online backend path updates \(\mathcal{M}_{cur}\) as a rolling buffer and builds \(\mathcal{M}_{st}\) from temporal continuity and cross-frame changes, while costly operations such as ASR backfilling, event refinement, indexing, and semantic extraction run asynchronously.

Query-time cost is also reduced.
Stable events are consolidated into a event-centric long-term memory, where multimodal evidence is pre-aligned under event anchors.
User-facing inference therefore retrieves compact event records rather than raw streams or fragmented modality-specific evidence.
A query router selects the cheapest sufficient source: \(\mathcal{M}_{cur}\) for current-scene queries, \(\mathcal{M}_{st}\) for recent recall, and \(\mathcal{M}_{lt}\) for retrospective or routine-level questions.
This keeps wearable interaction responsive while slower memory updates run in the background.

\subsection{Experience Retrieval and QA}

Given a user query \(q\), LightMem-Ego first routes the query to appropriate memory sources and then generates an answer from retrieved evidence. 
Formally, retrieval produces a query-specific evidence set \(\mathcal{R}(q)\subseteq \mathcal{M}_{cur}\cup\mathcal{M}_{st}\cup\mathcal{M}_{lt}\). 
The routing process is guided by the temporal scope and semantic intent of the query, naturally supporting time-aware, person-aware, place-aware, and event-aware retrieval. Retrieved evidence may include recent observations, short-term event records, episodic entries, semantic summaries, timestamps, representative frames, and transcript snippets.
These heterogeneous signals are fused into a compact evidence view \(E_q\), and the final answer is generated as \(\hat{y}=f(q,E_q)\). 
By making retrieval explicit and memory-grounded, LightMem-Ego supports both immediate questions about the present scene and retrospective questions about who, where, when, and what happened in the user’s past experience.

\begin{figure*}[t]
     \centering
     \includegraphics[width=\textwidth]{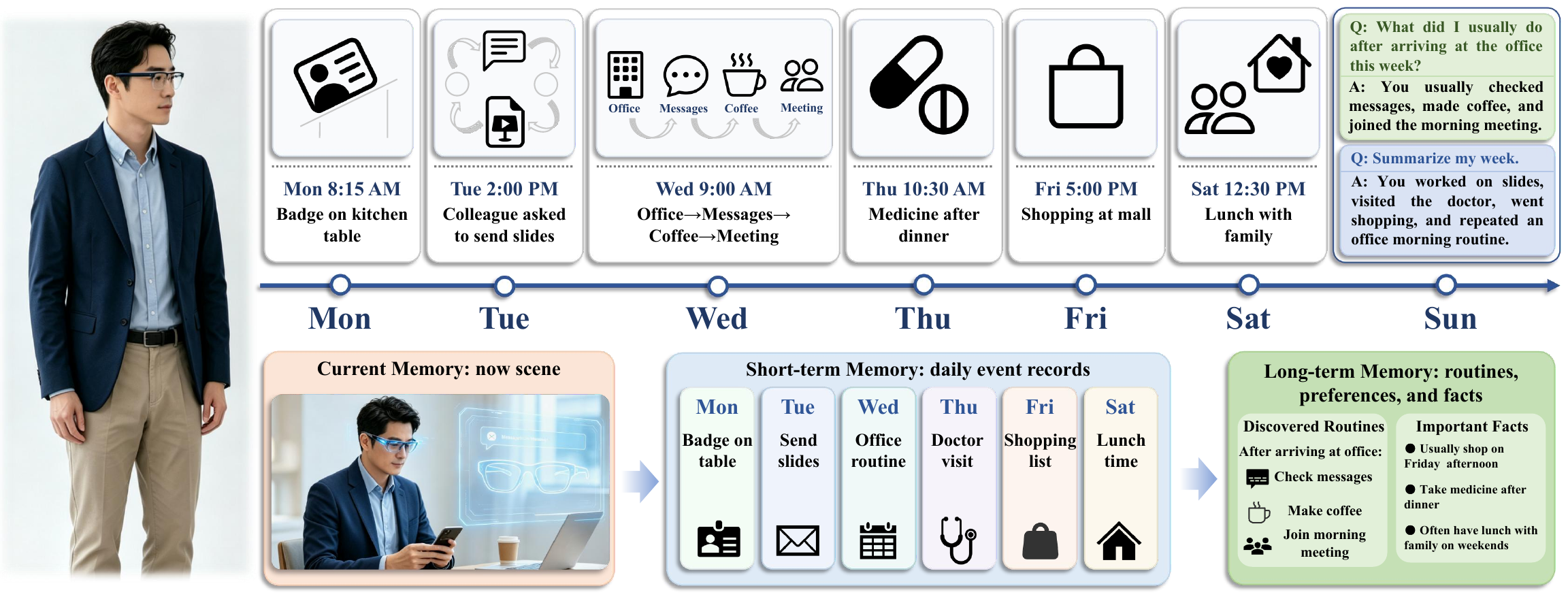}
     \caption{
     Representative demonstration scenarios of \textbf{LightMem-Ego}. 
     The system supports immediate assistance, conversation recall, life summarization, 
     and routine discovery by retrieving evidence from hierarchical memory.
     }
     \label{fig:demo_scenarios}
     \vspace{-5mm}
\end{figure*}

\section{Demonstration}

Figure~\ref{fig:demo_scenarios} summarizes the main demonstration scenarios of LightMem-Ego. 
We organize the demo around everyday memory needs that span 
immediate scene understanding, recent event recall, and long-term routine reasoning.

\paragraph{Immediate Assistance.}
LightMem-Ego supports immediate assistance and short-horizon recall in mobile and AI-glasses settings. 
For example, users may ask \emph{Where did I leave my badge?}, \emph{When did I last hold my keys?}, or \emph{What am I looking at now?} 
In these cases, the system retrieves relevant recent observations and event segments, and returns answers grounded in temporally localized memory evidence. 
This case demonstrates how online perception and short-term memory can work together in everyday environments.

\paragraph{Conversation Recall.}
LightMem-Ego also supports recall of recent spoken interactions. 
After a meeting or conversation, users can ask questions such as \emph{What did the doctor tell me after checking the report?} or \emph{What did my colleague ask me to do?} 
The system combines short-term event memory with aligned transcript context to recover the relevant conversational episode. 
This shows the value of multimodal memory for recovering spoken information beyond one-shot speech transcription.

\paragraph{Life Summarization and Routine Discovery.}
Beyond recovering isolated facts, LightMem-Ego can summarize recent experiences and reason over repeated patterns. 
A user may ask \emph{What did I do this afternoon?}, \emph{Summarize what happened during lunch}, or \emph{What do I usually do after arriving at the office?} 
The system aggregates event-centered evidence across multiple segments and, when appropriate, draws on long-term semantic memory to identify recurring activities, preferences, and routines. 
This scenario illustrates the ability of LightMem-Ego to move from event recall to higher-level summarization and habit-level reasoning.

\section{Quantitative Evaluation}

\subsection{Evaluation Setup}

We evaluate LightMem-Ego on three daily memory scenarios:
object finding, conversation recall, and life summarization. 
For each scenario, we construct queries with manually annotated gold
evidence. 
We report retrieval performance with Recall@1/3/5 and
MRR, QA performance with LLM- and human-judged accuracy, and
interactive latency with P50 / P90 retrieval, generation, and
end-to-end QA time on phone and glasses-style clients.

\subsection{Curated Evaluation Set}
The retrieval and QA results below use a small, intentionally balanced evaluation set for exercising the demonstration's three workflows, rather than for broad benchmarking.  
It contains 27 queries (nine per scenario) over five everyday-life videos totaling approximately 45.7 minutes; each query has a manually prepared answer and temporal gold-evidence intervals.  
We report per-scenario results as descriptive measurements of the current prototype.  
A larger streaming first-person benchmark with longer continuous recordings, cross-session queries, and delayed-memory conditions is under development and beyond the scope of this paper.

\begin{table}[ht]
\centering
\small
\setlength{\tabcolsep}{3pt}
\renewcommand{\arraystretch}{1.10}
\begin{tabular*}{\columnwidth}{@{\extracolsep{\fill}}lcccc@{}}
\toprule
\textbf{Scenario} & 
\textbf{R@1} & 
\textbf{R@3} & 
\textbf{R@5} & 
\textbf{MRR} \\
\midrule
Object Finding       & 22.2 & 66.7 & 77.8 & 0.454 \\
Conversation Recall  & 44.4 & 55.6 & 55.6 & 0.481 \\
Life Summarization   & 88.9 & 100.0 & 100.0 & 0.944 \\
\midrule
\textbf{Overall}     & \textbf{51.9} & \textbf{74.1} & \textbf{77.8} & \textbf{0.627} \\
\bottomrule
\end{tabular*}
\caption{
Memory retrieval accuracy across everyday-life scenarios.
R@k denotes Recall@k, and MRR measures the reciprocal rank of the first relevant memory entry.
}
\label{tab:retrieval_accuracy}
\vspace{-3mm}
\end{table}
\subsection{Memory Retrieval Accuracy}
Table~\ref{tab:retrieval_accuracy} summarizes retrieval accuracy in the three demonstration scenarios. 
LightMem-Ego retrieves relevant memory evidence within the top-3 results for most queries, achieving 74.1 overall R@3 and 0.627 MRR. 
The results suggest that the hierarchical memory store provides sufficiently accurate evidence selection for downstream experience QA in everyday object finding, conversation recall, and life summarization settings.

\begin{table}[ht]
\centering
\small
\setlength{\tabcolsep}{3pt}
\renewcommand{\arraystretch}{1.10}
\begin{tabular*}{\columnwidth}{@{\extracolsep{\fill}}lcc@{}}
\toprule
\textbf{Scenario} & 
\textbf{LLM-Judge} & 
\textbf{Human Acc.} \\
\midrule
Object Finding       & 44.4 & 55.6 \\
Conversation Recall  & 33.3 & 33.3 \\
Life Summarization   & 77.8 & 77.8 \\
\midrule
\textbf{Overall}     & \textbf{51.9} & \textbf{55.6} \\
\bottomrule
\end{tabular*}
\caption{
Experience QA accuracy on daily scenarios.
}
\label{tab:qa_accuracy}
\vspace{-3mm}
\end{table}
\subsection{Experience Question Answering Accuracy}
Table~\ref{tab:qa_accuracy} reports QA accuracy against annotated
experience evidence. 
LightMem-Ego obtains 51.9 LLM-judged accuracy and 55.6 human-judged accuracy overall, with the best performance on life summarization.
These results indicate that, when relevant evidence is retrieved, the system can produce usable memory-grounded answers for the demonstrated everyday QA scenarios, while fine-grained object finding and conversation recall remain more challenging in the current prototype.

\subsection{Latency on Mobile and Wearable Clients}
Table~\ref{tab:latency} reports the runtime of LightMem-Ego on
phone and glasses-style clients. 
For short-term memory QA, the system achieves interactive response times, with P50 end-to-end latency of 6.42s on phone and 6.95s on the glasses-style client.
Long-term memory QA requires additional retrieval and evidence
aggregation, leading to higher P50 latency of 10.57s and 8.61s,
respectively. 
This latency profile is consistent with the intended usage of the demo: recent-memory queries support near-interactive assistance, while long-term queries remain practical for retrospective recall, daily reflection, and life summaries.
\begin{table}[ht]
\centering
\small
\setlength{\tabcolsep}{3pt}
\renewcommand{\arraystretch}{1.10}
\begin{tabular*}{\columnwidth}{@{\extracolsep{\fill}}lcccc@{}}
\toprule
\multirow{2}{*}{\textbf{Stage}} &
\multicolumn{2}{c}{\textbf{Phone}} &
\multicolumn{2}{c}{\textbf{Glasses-style}} \\
\cmidrule(lr){2-3}
\cmidrule(lr){4-5}
&
\textbf{P50} & \textbf{P90} &
\textbf{P50} & \textbf{P90} \\
\midrule

\multicolumn{5}{@{}l}{\textit{\textbf{Short-term memory QA}}} \\
Retrieval       & 76 ms & 131 ms & 44 ms & 87 ms \\
TTFT (end-to-end)& 532 ms & 643 ms & 423 ms & 494 ms \\
Answer gen.     & 6.13 s  & 10.11 s  & 6.81 s  & 9.14 s  \\
End-to-end QA   & 6.42 s  & 10.34 s  & 6.95 s  & 9.31 s  \\

\midrule

\multicolumn{5}{@{}l}{\textit{\textbf{Long-term memory QA}}} \\
Retrieval       & 2.99 s & 3.84 s & 3.06 s  & 3.44 s  \\
TTFT (end-to-end)& 4.64 s  & 5.44 s  & 4.74 s & 5.07 s  \\
Answer gen.     & 5.78 s & 9.56 s & 4.37 s  & 9.16 s  \\
End-to-end QA   & 10.57 s & 13.93 s & 8.61 s  & 13.60 s  \\

\bottomrule
\end{tabular*}
\caption{
Latency breakdown by memory scope and client.
P50 / P90 are 50th / 90th percentile latencies.
}
\label{tab:latency}
\vspace{-3mm}
\end{table}

% Optional symbols
\newcommand{\cmark}{\(\checkmark\)}
\newcommand{\xmark}{---}
\newcommand{\pmark}{\emph{Partial}}

\begin{table*}[ht]
\centering
\scriptsize
\setlength{\tabcolsep}{2.5pt}
\renewcommand{\arraystretch}{1.15}
\begin{tabularx}{\textwidth}{@{}
>{\raggedright\arraybackslash}p{0.18\textwidth}
>{\raggedright\arraybackslash}p{0.20\textwidth}
*{5}{>{\centering\arraybackslash}X}
@{}}
\toprule
\textbf{System} &
\textbf{Platform and Input} &
\makecell{\textbf{Real-time}\\\textbf{Visual-Audio}\\\textbf{Stream}} &
\makecell{\textbf{Current /}\\\textbf{Short-term}\\\textbf{MM Memory}} &
\makecell{\textbf{Long-term}\\\textbf{MM Episodic}\\\textbf{Memory}} &
\makecell{\textbf{Long-term}\\\textbf{Semantic}\\\textbf{Memory}} &
\makecell{\textbf{Timestamped}\\\textbf{Evidence}\\\textbf{Retrieval}} \\
\midrule

ChatGPT Memory &
Text chat &
\xmark &
\xmark &
\xmark &
\pmark &
\xmark \\

Mem0-style memory &
Text and agent memory &
\xmark &
\pmark &
\xmark &
\cmark &
\pmark \\

Memories.ai &
Video archives and visual memory platform &
\pmark &
\pmark &
\cmark &
\pmark &
\pmark \\

Gemini Live &
Phone &
\cmark &
\pmark &
\xmark &
\xmark &
\xmark \\

Ray-Ban Meta AI Glasses &
Glasses &
\pmark &
\pmark &
\xmark &
\xmark &
\xmark \\

Vinci &
Phone or wearable camera &
\cmark &
\cmark &
\pmark &
\pmark &
\pmark \\

VisualClaw &
Streaming video with agent workspace &
\pmark &
\pmark &
\xmark &
\pmark &
\pmark \\

VisionClaw &
Smart glasses &
\cmark &
\pmark &
\xmark &
\xmark &
\pmark \\

Egocentric Co-Pilot &
Smart glasses with web agents &
\cmark &
\cmark &
\pmark &
\pmark &
\pmark \\

EgoButler &
AI-glasses egocentric video and audio &
\pmark &
\pmark &
\pmark &
\pmark &
\cmark \\

\textbf{LightMem-Ego} &
\textbf{Phone and glasses-style client} &
\cmark &
\cmark &
\cmark &
\cmark &
\cmark \\

\bottomrule
\end{tabularx}
\vspace{-2mm}
\caption{
Capability comparison with representative commercial assistants, text-based memory systems, and egocentric multimodal assistants.
The table compares publicly described system capabilities rather than experimentally measured performance.
We mark a capability as supported only when it is implemented as an explicit first-class component.
\emph{Partial} indicates limited, implicit, offline, session-level, or modality-restricted support;
--- indicates that the capability is not explicitly supported or not publicly described.
}
\label{tab:capability_comparison}
\vspace{-5mm}
\end{table*}

\subsection{Capability Comparison with Existing Assistants and Memory Systems}
We further compare LightMem-Ego with representative commercial assistants, text-based memory systems, and recent egocentric multimodal assistants in terms of publicly described capabilities. 
As shown in Table~\ref{tab:capability_comparison}, existing systems cover different parts of the design space: ChatGPT Memory and Mem0-style systems mainly focus on conversational or text-based agent memory, while Memories.ai explores long-term visual memory for searchable video archives~\cite{GPT-4o, GPT-5, Mem0, MemoriesAI}. Gemini Live and Ray-Ban Meta AI Glasses support real-time multimodal interaction, while recent egocentric assistants such as Vinci, VisualClaw, VisionClaw, Egocentric Co-Pilot, and EgoButler explore real-time perception, wearable assistance, tool use, or long-context egocentric QA~\cite{Vinci, VisualClaw, VisionClaw, Egocentric_Co-Pilot, EgoLife}.
However, these systems generally do not expose an explicit hierarchy that jointly supports current multimodal memory, short-term event memory, long-term multimodal episodic memory, semantic routine memory, and timestamped evidence retrieval.

In contrast, LightMem-Ego is designed specifically for everyday-life memory assistance over streaming visual-audio experience.
Rather than emphasizing only in-the-moment perception, text-based personalization, or agentic task execution, it organizes user experience into current, short-term, and long-term memory stores, enabling both immediate assistance and retrospective experience QA such as object finding, conversation recall, life summarization, and routine discovery.

\section{Conclusion}

We introduced LightMem-Ego, a streaming multimodal memory system for everyday-life assistance. 
The system connects mobile and wearable multimodal capture with hierarchical memory construction and memory-grounded question answering, enabling support for object finding, conversation recall, life summarization, and routine understanding within a single backend. 
Our demonstration and evaluation suggest that explicit long-horizon multimodal memory is a promising systems direction for personal AI assistants.

Ultimately, building real-world personal AI assistants requires moving beyond language understanding alone toward the ability to capture, organize, and recall human experience over time. 
LightMem-Ego takes a step in this direction by showing how multimodal memory can support AI systems that not only understand utterances, but also understand lived experience.

\section*{Limitations}

LightMem-Ego remains constrained by several system-level factors. Its current implementation relies on upstream API calls for speech recognition, visual understanding, language generation, and retrieval-based question answering, making both accuracy and latency sensitive to external model reliability, runtime variation, and rate limits. 
Errors in transcripts, visual descriptions, or timestamp alignment may further propagate into memory construction and affect answer faithfulness. 
In addition, maintaining multimodal memory over continuous egocentric streams still incurs non-trivial overhead, as the system repeatedly performs segmentation, summarization, embedding, indexing, and storage.
The memory update mechanism is also preliminary: it lacks a principled policy for revising, merging, forgetting, or promoting memories. 
Future work will improve efficiency, robustness, and adaptive memory lifecycle management for longer and more diverse deployments.

\section*{Ethical and Privacy Considerations}

LightMem-Ego relies on egocentric visual and audio streams, which may contain sensitive information about users and nearby bystanders, including private conversations, faces, documents, locations, screens, and daily routines. 
Because the system converts these streams into persistent and queryable memory, privacy risks can arise from both raw data and derived representations such as transcripts, event summaries, embeddings, and long-term semantic memories.

The current prototype is intended as a research demonstration and has not yet implemented a complete privacy-preserving pipeline. 
It does not automatically redact sensitive content, manage bystander consent, enforce fine-grained access control, or provide mature retention and deletion policies. 
Future work will integrate privacy protection into the memory lifecycle, including on-device preprocessing, selective capture, sensitive-content filtering, encrypted storage, user-controlled memory editing and deletion, and privacy-aware consolidation that avoids promoting incidental sensitive information into long-term memory.

\section*{Acknowledgements}
We would like to express our sincere gratitude to the anonymous reviewers for their thoughtful and constructive feedback. This work was supported by the National Natural Science Foundation of China (No. 62576307), and Information Technology Center and State Key Lab of CAD\&CG, Zhejiang University. This work was sponsored by CCF-Lenovo Blue Ocean Research Fund.

% \input{section/background}

% \input{section/data_construction}

% \input{section/experiment}

% \input{section/analysis}

% \input{section/agent}

% \input{section/related_work}

% \input{section/conclusion}

% \clearpage

% \input{section/limitation}

\bibliography{custom}

\begin{thebibliography}{37}
\providecommand{\natexlab}[1]{#1}

\bibitem[{Alam et~al.(2026)Alam, Siam, Proulx, Fort, Newcombe, Kim, and Zhang}]{SuperMemory-VQA}
Samiul Alam, Shakhrul~Iman Siam, Michael~J Proulx, James Fort, Richard Newcombe, Hyo~Jin Kim, and Mi~Zhang. 2026.
\newblock \href {https://arxiv.org/abs/2606.00825} {Supermemory-vqa: An egocentric visual question-answering benchmark for long-horizon memory}.
\newblock \emph{arXiv preprint arXiv:2606.00825}.

\bibitem[{Chen et~al.(2026{\natexlab{a}})Chen, Zheng, Li, Xiao, Xu, Qiao, Fang, Deng, Yao, Wang, Zhang, Li, Chen, and Deng}]{EM²Mem}
Yijun Chen, Yaqi Zheng, Yanya Li, Boyi Xiao, Buqiang Xu, Shuofei Qiao, Jizhan Fang, Xinle Deng, Yunzhi Yao, Xuehai Wang, Liuxin Zhang, Hui Li, Huajun Chen, and Shumin Deng. 2026{\natexlab{a}}.
\newblock \href {https://doi.org/10.18653/2026.findings-emnlp.903} {{EM}{\texttwosuperior}{M}em: Event-centric multimodal memory for large language models}.
\newblock In \emph{Findings of the {A}ssociation for {C}omputational {L}inguistics: {EMNLP} 2026}, pages 19117--19144, Budapest, Hungary. Association for Computational Linguistics.

\bibitem[{Chen et~al.(2026{\natexlab{b}})Chen, Zheng, Li, Xiao, Xu, Qiao, Fang, Deng, Yao, Wang et~al.}]{EM2Mem}
Yijun Chen, Yaqi Zheng, Yanya Li, Boyi Xiao, Buqiang Xu, Shuofei Qiao, Jizhan Fang, Xinle Deng, Yunzhi Yao, Xuehai Wang, et~al. 2026{\natexlab{b}}.
\newblock \href {https://arxiv.org/abs/2609.00551} {Em\^{} 2mem: Event-centric multimodal memory for large language models}.
\newblock \emph{arXiv preprint arXiv:2609.00551}.

\bibitem[{Chhikara et~al.(2025)Chhikara, Khant, Aryan, Singh, and Yadav}]{Mem0}
Prateek Chhikara, Dev Khant, Saket Aryan, Taranjeet Singh, and Deshraj Yadav. 2025.
\newblock \href {https://doi.org/10.3233/FAIA251160} {Mem0: Building production-ready {AI} agents with scalable long-term memory}.
\newblock In \emph{{ECAI} 2025 - 28th European Conference on Artificial Intelligence, 25-30 October 2025, Bologna, Italy - Including 14th Conference on Prestigious Applications of Intelligent Systems {(PAIS} 2025)}, volume 413 of \emph{Frontiers in Artificial Intelligence and Applications}, pages 2993--3000. {IOS} Press.

\bibitem[{Deng et~al.(2026{\natexlab{a}})Deng, Xue, Chen, Mao, Zhong, Xu, Fang, Xu, Wu, Xu et~al.}]{MobileMem}
Xinle Deng, Yida Xue, Yijun Chen, Mingjun Mao, Ruobin Zhong, Buqiang Xu, Jizhan Fang, Haoming Xu, Tingwei Wu, Yajing Xu, et~al. 2026{\natexlab{a}}.
\newblock \href {https://iclr.cc/virtual/2026/10012468} {Mobilemem: Evaluating long-horizon memory for language agents in real-world mobile environments}.
\newblock In \emph{ICLR 2026 Workshop on Lifelong Agents: Learning, Aligning, Evolving}.

\bibitem[{Deng et~al.(2026{\natexlab{b}})Deng, Xue, Ru, Xu, Qiao, Wang, Chen, Xu, Jiang, Jiang et~al.}]{deng2026mobilemem}
Xinle Deng, Yida Xue, Xiangyuan Ru, Haoming Xu, Shuofei Qiao, Mengru Wang, Yijun Chen, Buqiang Xu, Chen Jiang, Yuchen~Eleanor Jiang, et~al. 2026{\natexlab{b}}.
\newblock Mobilemem: Learning from a year of mobile experiences.
\newblock \emph{arXiv preprint arXiv:2608.13606}.

\bibitem[{Du(2026)}]{Memory_for_Autonomous_LLM_Agents}
Pengfei Du. 2026.
\newblock \href {https://doi.org/10.48550/ARXIV.2603.07670} {Memory for autonomous {LLM} agents:mechanisms, evaluation, and emerging frontiers}.
\newblock \emph{CoRR}, abs/2603.07670.

\bibitem[{Fan et~al.(2024)Fan, Ma, Wu, Du, Li, Gao, and Li}]{VideoAgent}
Yue Fan, Xiaojian Ma, Rujie Wu, Yuntao Du, Jiaqi Li, Zhi Gao, and Qing Li. 2024.
\newblock \href {https://doi.org/10.1007/978-3-031-72670-5\_5} {Videoagent: {A} memory-augmented multimodal agent for video understanding}.
\newblock In \emph{Computer Vision - {ECCV} 2024 - 18th European Conference, Milan, Italy, September 29-October 4, 2024, Proceedings, Part {XXII}}, Lecture Notes in Computer Science, pages 75--92. Springer.

\bibitem[{Fang et~al.(2025)Fang, Deng, Xu, Jiang, Tang, Xu, Deng, Yao, Wang, Qiao, Chen, and Zhang}]{LightMem}
Jizhan Fang, Xinle Deng, Haoming Xu, Ziyan Jiang, Yuqi Tang, Ziwen Xu, Shumin Deng, Yunzhi Yao, Mengru Wang, Shuofei Qiao, Huajun Chen, and Ningyu Zhang. 2025.
\newblock \href {https://doi.org/10.48550/ARXIV.2510.18866} {Lightmem: Lightweight and efficient memory-augmented generation}.
\newblock \emph{CoRR}, abs/2510.18866.

\bibitem[{Gao et~al.(2026)Gao, Zhang, Wang, Chen, Cao, Wang, Zhu, Min, Sun, Zhu, and Zhai}]{LifeEval}
Hengjian Gao, Kaiwei Zhang, Shibo Wang, Mingjie Chen, Qihang Cao, Xianfeng Wang, Yucheng Zhu, Xiongkuo Min, Wei Sun, Dandan Zhu, and Guangtao Zhai. 2026.
\newblock \href {https://doi.org/10.48550/ARXIV.2603.00490} {Lifeeval: {A} multimodal benchmark for assistive {AI} in egocentric daily life tasks}.
\newblock \emph{CoRR}, abs/2603.00490.

\bibitem[{{Gemini Team}(2025)}]{Gemini-2.5-pro}
{Gemini Team}. 2025.
\newblock \href {https://doi.org/10.48550/ARXIV.2507.06261} {Gemini 2.5: Pushing the frontier with advanced reasoning, multimodality, long context, and next generation agentic capabilities}.
\newblock \emph{CoRR}, abs/2507.06261.

\bibitem[{Huang et~al.(2025{\natexlab{a}})Huang, Xu, Pei, He, Chen, Zhang, Yang, Nie, Liu, Fan, Lin, Fang, Li, Yuan, Chen, Wang, Wang, Qiao, and Wang}]{Vinci_new}
Yifei Huang, Jilan Xu, Baoqi Pei, Yuping He, Guo Chen, Mingfang Zhang, Lijin Yang, Zheng Nie, Jinyao Liu, Guoshun Fan, Dechen Lin, Fang Fang, Kunpeng Li, Chang Yuan, Xinyuan Chen, Yaohui Wang, Yali Wang, Yu~Qiao, and Limin Wang. 2025{\natexlab{a}}.
\newblock \href {https://doi.org/10.48550/ARXIV.2503.04250} {An egocentric vision-language model based portable real-time smart assistant}.
\newblock \emph{CoRR}, abs/2503.04250.

\bibitem[{Huang et~al.(2025{\natexlab{b}})Huang, Xu, Pei, Yang, Zhang, He, Chen, Chen, Wang, Nie, Liu, Lin, Fang, Li, Yuan, Qiao, Wang, and Wang}]{Vinci}
Yifei Huang, Jilan Xu, Baoqi Pei, Lijin Yang, Mingfang Zhang, Yuping He, Guo Chen, Xinyuan Chen, Yaohui Wang, Zheng Nie, Jinyao Liu, Dechen Lin, Fang Fang, Kunpeng Li, Chang Yuan, Yu~Qiao, Yali Wang, and Limin Wang. 2025{\natexlab{b}}.
\newblock \href {https://doi.org/10.1145/3749513} {Vinci: {A} real-time smart assistant based on egocentric vision-language model for portable devices}.
\newblock \emph{Proc. {ACM} Interact. Mob. Wearable Ubiquitous Technol.}, 9(3):88:1--88:33.

\bibitem[{Jiang et~al.(2026)Jiang, Yuan, Qu, Lin, Liu, Fan, and Li}]{SUPERGLASSES}
Zhuohang Jiang, Xu~Yuan, Haohao Qu, Shanru Lin, Kanglong Liu, Wenqi Fan, and Qing Li. 2026.
\newblock \href {https://doi.org/10.48550/ARXIV.2602.22683} {{SUPERGLASSES:} benchmarking vision language models as intelligent agents for {AI} smart glasses}.
\newblock \emph{CoRR}, abs/2602.22683.

\bibitem[{Li et~al.(2025)Li, Song, Xi, Wang, Tang, Niu, Chen, Yang, Li, Yu, Zhao, Wang, Liu, Lin, Wang, Huo, Chen, Chen, Li, Tao, Ren, Lai, Wu, Tang, Wang, Fan, Zhang, Zhang, Yan, Yang, Xu, Xu, Chen, Wang, Yang, Zhang, Xu, Chen, and Xiong}]{MemOS}
Zhiyu Li, Shichao Song, Chenyang Xi, Hanyu Wang, Chen Tang, Simin Niu, Ding Chen, Jiawei Yang, Chunyu Li, Qingchen Yu, Jihao Zhao, Yezhaohui Wang, Peng Liu, Zehao Lin, Pengyuan Wang, Jiahao Huo, Tianyi Chen, Kai Chen, Kehang Li, Zhen Tao, Junpeng Ren, Huayi Lai, Hao Wu, Bo~Tang, Zhenren Wang, Zhaoxin Fan, Ningyu Zhang, Linfeng Zhang, Junchi Yan, Mingchuan Yang, Tong Xu, Wei Xu, Huajun Chen, Haofeng Wang, Hongkang Yang, Wentao Zhang, Zhi{-}Qin~John Xu, Siheng Chen, and Feiyu Xiong. 2025.
\newblock \href {https://doi.org/10.48550/ARXIV.2507.03724} {Memos: {A} memory {OS} for {AI} system}.
\newblock \emph{CoRR}, abs/2507.03724.

\bibitem[{Liu et~al.(2026)Liu, Lee, Gonzalez, Gonz{\'{a}}lez{-}Franco, and Suzuki}]{VisionClaw}
Xiaoan Liu, Daeho Lee, Eric~J. Gonzalez, Mar Gonz{\'{a}}lez{-}Franco, and Ryo Suzuki. 2026.
\newblock \href {https://doi.org/10.48550/ARXIV.2604.03486} {Visionclaw: Always-on {AI} agents through smart glasses}.
\newblock \emph{CoRR}, abs/2604.03486.

\bibitem[{Long et~al.(2025)Long, He, Ye, Pan, Lin, Li, Zhao, and Li}]{M3-Agent}
Lin Long, Yichen He, Wentao Ye, Yiyuan Pan, Yuan Lin, Hang Li, Junbo Zhao, and Wei Li. 2025.
\newblock \href {https://doi.org/10.48550/ARXIV.2508.09736} {Seeing, listening, remembering, and reasoning: {A} multimodal agent with long-term memory}.
\newblock \emph{CoRR}, abs/2508.09736.

\bibitem[{{Memories.ai}(2026)}]{MemoriesAI}
{Memories.ai}. 2026.
\newblock \href {https://memories.ai/} {Memories.ai: Ai video analysis and visual memory platform}.
\newblock Accessed: 2026-07-11.

\bibitem[{OpenAI(2024)}]{GPT-4o}
OpenAI. 2024.
\newblock \href {https://doi.org/10.48550/ARXIV.2410.21276} {Gpt-4o system card}.
\newblock \emph{CoRR}, abs/2410.21276.

\bibitem[{OpenAI(2026)}]{GPT-5}
OpenAI. 2026.
\newblock \href {https://doi.org/10.48550/ARXIV.2601.03267} {Openai {GPT-5} system card}.
\newblock \emph{CoRR}, abs/2601.03267.

\bibitem[{Packer et~al.(2023)Packer, Fang, Patil, Lin, Wooders, and Gonzalez}]{MemGPT}
Charles Packer, Vivian Fang, Shishir~G. Patil, Kevin Lin, Sarah Wooders, and Joseph~E. Gonzalez. 2023.
\newblock \href {https://doi.org/10.48550/ARXIV.2310.08560} {Memgpt: Towards llms as operating systems}.
\newblock \emph{CoRR}, abs/2310.08560.

\bibitem[{Pu et~al.(2025)Pu, Zhang, Sendhilnathan, Freitag, Sodhi, and Jonker}]{ProMemAssist}
Kevin Pu, Ting Zhang, Naveen Sendhilnathan, Sebastian Freitag, Raj Sodhi, and Tanya~R. Jonker. 2025.
\newblock \href {https://doi.org/10.1145/3746059.3747770} {Promemassist: Exploring timely proactive assistance through working memory modeling in multi-modal wearable devices}.
\newblock In \emph{Proceedings of the 38th Annual {ACM} Symposium on User Interface Software and Technology, {UIST} 2025, Busan, Korea, 28 September 2025 - 1 October 2025}, pages 56:1--56:19. {ACM}.

\bibitem[{Qiao et~al.(2026)Qiao, Pan, Chen, Feng, Wu, Fang, Zhu, Deng, and Yilmaz}]{qiao2026delta}
Shuofei Qiao, Xuanguang Pan, Yijun Chen, Wendi Feng, Bin Wu, Runnan Fang, Yuqi Zhu, Shumin Deng, and Emine Yilmaz. 2026.
\newblock \href {https://arxiv.org/abs/2610.34238} {{$\Delta\tau$-Bench: Pushing Agents Toward a Moving Capability}}.
\newblock \emph{arXiv preprint arXiv:2610.34238}.

\bibitem[{Tang et~al.(2026)Tang, Zhang, Qin, Yu, Qiu, Gou, Xiong, Qingwei, Rajmohan, Zhang, and Wu}]{EgoMemory}
Yuanmin Tang, Jue Zhang, Xiaoting Qin, Jing Yu, Meikang Qiu, Gaopeng Gou, Gang Xiong, Lin Qingwei, Saravan Rajmohan, Dongmei Zhang, and Qi~Wu. 2026.
\newblock \href {https://www.microsoft.com/en-us/research/publication/egomemory-memory-augmented-personalized-retrieval-for-long-context-egocentric-video/} {Egomemory: Memory-augmented personalized retrieval for long-context egocentric video}.
\newblock In \emph{ACL Findings}.

\bibitem[{Tian et~al.(2025)Tian, Wang, Guo, Wu, Dong, Wang, Yang, Zhang, Zhu, and Liu}]{Ego-R1}
Shulin Tian, Ruiqi Wang, Hongming Guo, Penghao Wu, Yuhao Dong, Xiuying Wang, Jingkang Yang, Hao Zhang, Hongyuan Zhu, and Ziwei Liu. 2025.
\newblock \href {https://doi.org/10.48550/ARXIV.2506.13654} {Ego-r1: Chain-of-tool-thought for ultra-long egocentric video reasoning}.
\newblock \emph{CoRR}, abs/2506.13654.

\bibitem[{Tran et~al.(2025)Tran, Bailer, Dang{-}Nguyen, Healy, Hodges, J{\'{o}}nsson, Rossetto, Schoeffmann, Tran, Vadicamo, and Gurrin}]{Lifelong_Retrieval}
Allie Tran, Werner Bailer, Duc{-}Tien Dang{-}Nguyen, Graham Healy, Steve Hodges, Bj{\"{o}}rn~{\TH}{\'{o}}r J{\'{o}}nsson, Luca Rossetto, Klaus Schoeffmann, Minh{-}Triet Tran, Lucia Vadicamo, and Cathal Gurrin. 2025.
\newblock \href {https://doi.org/10.48550/ARXIV.2506.06743} {The state-of-the-art in lifelog retrieval: {A} review of progress at the {ACM} lifelog search challenge workshop 2022-24}.
\newblock \emph{CoRR}, abs/2506.06743.

\bibitem[{Tu et~al.(2026)Tu, Chen, Wang, Han, Wu, Chen, Ji, Xiong, Liu, Xia et~al.}]{VisualClaw}
Haoqin Tu, Jianwen Chen, Zijun Wang, Siwei Han, Juncheng Wu, Hardy Chen, Haonian Ji, Kaiwen Xiong, Jiaqi Liu, Peng Xia, et~al. 2026.
\newblock \href {https://arxiv.org/abs/2606.16295} {Visualclaw: A real-time, personalized agent for the physical world}.
\newblock \emph{arXiv preprint arXiv:2606.16295}.

\bibitem[{Wang et~al.(2026{\natexlab{a}})Wang, Qin, Liu, Li, Xu, Zhang, Chen, Xue, and Deng}]{wang2026learning}
Xuehai Wang, Haowei Qin, Tongxin Liu, Junkai Li, Buqiang Xu, Jintian Zhang, Yijun Chen, Zirui Xue, and Shumin Deng. 2026{\natexlab{a}}.
\newblock \href {https://arxiv.org/abs/2608.31076} {Learning to evaluate before improving: Automatic rubric induction for automatic research agents}.
\newblock \emph{arXiv preprint arXiv:2608.31076}.

\bibitem[{Wang et~al.(2026{\natexlab{b}})Wang, Zhang, Yu, Zhang, Zhao, Yoon, Lee, Bertasius, and Bansal}]{EgoMemReason}
Ziyang Wang, Yue Zhang, Shoubin Yu, Ce~Zhang, Zengqi Zhao, Jaehong Yoon, Hyunji Lee, Gedas Bertasius, and Mohit Bansal. 2026{\natexlab{b}}.
\newblock \href {https://doi.org/10.48550/ARXIV.2605.09874} {Egomemreason: {A} memory-driven reasoning benchmark for long-horizon egocentric video understanding}.
\newblock \emph{CoRR}, abs/2605.09874.

\bibitem[{Xiao et~al.(2026)Xiao, Zhang, Zhu, and Yao}]{MyEgo}
Junbin Xiao, Shenglang Zhang, Pengxiang Zhu, and Angela Yao. 2026.
\newblock \href {https://doi.org/10.48550/ARXIV.2604.01966} {Ego-grounding for personalized question-answering in egocentric videos}.
\newblock \emph{CoRR}, abs/2604.01966.

\bibitem[{Xu et~al.(2026{\natexlab{a}})Xu, Chen, Fang, Zhong, Yao, Zhu, Du, and Deng}]{StructMem}
Buqiang Xu, Yijun Chen, Jizhan Fang, Ruobin Zhong, Yunzhi Yao, Yuqi Zhu, Lun Du, and Shumin Deng. 2026{\natexlab{a}}.
\newblock \href {https://doi.org/10.48550/ARXIV.2604.21748} {Structmem: Structured memory for long-horizon behavior in llms}.
\newblock \emph{CoRR}, abs/2604.21748.

\bibitem[{Xu et~al.(2026{\natexlab{b}})Xu, Xue, Chen, Fu, Wu, Huang, Jiang, Fang, Deng, Chen et~al.}]{TokenPilot}
Buqiang Xu, Zirui Xue, Dianmou Chen, Chenyang Fu, Chiyu Wu, Caiying Huang, Chen Jiang, Jizhan Fang, Xinle Deng, Yijun Chen, et~al. 2026{\natexlab{b}}.
\newblock \href {https://arxiv.org/abs/2606.17016} {Tokenpilot: Cache-efficient context management for llm agents}.
\newblock \emph{arXiv preprint arXiv:2606.17016}.

\bibitem[{Xu et~al.(2025)Xu, Zeng, Tong, Li, and Srivastava}]{AutoLife}
Huatao Xu, Zilin Zeng, Panrong Tong, Mo~Li, and Mani~B. Srivastava. 2025.
\newblock \href {https://doi.org/10.1145/3770683} {Autolife: Automatic life journaling with smartphones and llms}.
\newblock \emph{Proc. {ACM} Interact. Mob. Wearable Ubiquitous Technol.}, 9(4):226:1--226:29.

\bibitem[{Yang et~al.(2025)Yang, Liu, Guo, Dong, Zhang, Zhang, Wang, Zhou, Xie, Wang, Ouyang, Lin, Cominelli, Cai, Li, Zhang, Zhang, Hong, Widmer, Gringoli, Yang, and Liu}]{EgoLife}
Jingkang Yang, Shuai Liu, Hongming Guo, Yuhao Dong, Xiamengwei Zhang, Sicheng Zhang, Pengyun Wang, Zitang Zhou, Binzhu Xie, Ziyue Wang, Bei Ouyang, Zhengyu Lin, Marco Cominelli, Zhongang Cai, Bo~Li, Yuanhan Zhang, Peiyuan Zhang, Fangzhou Hong, Joerg Widmer, Francesco Gringoli, Lei Yang, and Ziwei Liu. 2025.
\newblock \href {https://doi.org/10.1109/CVPR52734.2025.02690} {Egolife: Towards egocentric life assistant}.
\newblock In \emph{{IEEE/CVF} Conference on Computer Vision and Pattern Recognition, {CVPR} 2025, Nashville, TN, USA, June 11-15, 2025}, pages 28885--28900. Computer Vision Foundation / {IEEE}.

\bibitem[{Yang et~al.(2026)Yang, Huang, Cai, Sun, Fang, He, Xie, Deng, Zhang, Song, and Zhang}]{Egocentric_Co-Pilot}
Sicheng Yang, Yukai Huang, Weitong Cai, Shitong Sun, Fengyi Fang, You He, Yiqiao Xie, Jiankang Deng, Hang Zhang, Jifei Song, and Zhensong Zhang. 2026.
\newblock \href {https://doi.org/10.1145/3774904.3792996} {Egocentric co-pilot: Web-native smart-glasses agents for assistive egocentric {AI}}.
\newblock In \emph{Proceedings of the {ACM} Web Conference 2026, {WWW} 2026, Dubai, United Arab Emirates, originally scheduled for April 13-17, 2026, rescheduled for June 29 - July 3, 2026}, pages 8862--8873. {ACM}.

\bibitem[{Yeo et~al.(2025)Yeo, Kim, Yoon, and Hwang}]{WorldMM}
Woongyeong Yeo, Kangsan Kim, Jaehong Yoon, and Sung~Ju Hwang. 2025.
\newblock \href {https://doi.org/10.48550/ARXIV.2512.02425} {Worldmm: Dynamic multimodal memory agent for long video reasoning}.
\newblock \emph{CoRR}, abs/2512.02425.

\bibitem[{Zhou et~al.(2026)Zhou, Zhou, Han, Xu, Li, Li, Xiong, and Wu}]{Are_we_ready}
Wei Zhou, Xuanhe Zhou, Shaokun Han, Hongming Xu, Guoliang Li, Zhiyu Li, Feiyu Xiong, and Fan Wu. 2026.
\newblock \href {https://doi.org/10.48550/ARXIV.2606.24775} {Are we ready for an agent-native memory system?}
\newblock \emph{CoRR}, abs/2606.24775.

\end{thebibliography}
% \nocite{*}

\clearpage
% \onecolumn

\appendix
\section*{Appendix}
\section{Implementation Details}

LightMem-Ego is implemented as a session-oriented client--server system.
The phone and glasses-style clients remain lightweight, while the backend runs stream alignment, memory construction, retrieval, and answer generation as independent services.
Table~\ref{tab:implementation_details} summarizes the configuration used in our current prototype.

\subsection{Streaming Preprocessing}

The clients upload timestamped image frames and short audio chunks through a unified streaming interface.
The backend samples the visual stream at 1 frame per second (FPS) and groups audio into non-overlapping 5-second windows.
Audio is transcribed using the medium WhisperX model with FP16 inference on GPU; the current deployment supports Chinese and English word alignment.
Both modalities retain timestamps relative to the start of their session, allowing transcripts generated later to be attached to the corresponding visual interval.

The stream is materialized into fixed 30-second base segments for predictable incremental processing.
For each segment, we combine eight uniformly sampled frames with change-sensitive candidates detected from the 1-FPS stream using HSV-histogram and perceptual-hash differences, retaining at most 16 candidate keyframes.
Up to eight evenly selected keyframes are sent to an OpenAI-compatible multimodal model, which produces a structured event record containing a caption, scene, objects, actions, state changes, speakers, and confidence.
The change signal is used for keyframe selection rather than semantic boundary detection; transcripts and refined descriptions can be backfilled asynchronously when they become available.

\begin{table*}[t]
\centering
\scriptsize
\setlength{\tabcolsep}{3.5pt}
\renewcommand{\arraystretch}{1.13}
\begin{tabularx}{\textwidth}{@{}
>{\raggedright\arraybackslash}p{0.16\textwidth}
>{\raggedright\arraybackslash}p{0.36\textwidth}
>{\raggedright\arraybackslash}X
@{}}
\toprule
\textbf{Component} & \textbf{Prototype configuration} & \textbf{Purpose in LightMem-Ego} \\
\midrule
Visual stream & 1 FPS; session-relative timestamps & Low-bandwidth, temporally aligned egocentric capture \\
Audio and ASR & 5-s non-overlapping windows; WhisperX medium; GPU FP16; Chinese/English alignment & Incremental speech transcription and alignment with visual events \\
Base segmentation & Fixed 30-s intervals & Predictable units for online construction and asynchronous refinement \\
Keyframe selection & 8 uniform samples plus HSV-histogram/perceptual-hash change candidates; at most 16 candidates and 8 captioning frames & Compact coverage of both temporal extent and salient visual changes \\
Structured visual evidence & OpenAI-compatible multimodal API; deterministic decoding & Caption, scene, objects, actions, state changes, speakers, and confidence \\
Current memory & 30-s rolling window; 10-s core; at most 30 frames; 60-s stale threshold & Immediate scene and ongoing-activity questions \\
Short-term memory & Latest 30 min; at most 1,000 event records & Recent recall before long-term consolidation finishes \\
Long-term memory & Event-centric episodic and semantic stores & Retrospective event recall and higher-level personal knowledge \\
Embeddings & Qwen3-Embedding-4B (text); VLM2Vec-V2.0 (visual) & Cross-modal indexing and evidence retrieval \\
Routing and evidence & Deterministic, multi-store, readiness-aware routing and multimodal evidence packing & Select available memory sources and construct a compact grounded prompt \\
Remote answer profile & Configurable OpenAI-compatible API model & Flexible access to hosted answer models \\
Local answer profile & Qwen3.5-9B via vLLM; 16,384-token context; non-thinking mode; at most 4,096 output tokens & Avoid network and remote queueing overhead when lower TTFT is preferred \\
Runtime execution & Independent ingestion, ASR, visual, memory, and query workers & Keep interaction online while expensive consolidation runs asynchronously \\
\bottomrule
\end{tabularx}
\caption{
Implementation configuration of the LightMem-Ego prototype.
The answer model is deployment-selectable: the remote profile uses an OpenAI-compatible API, while the local profile redirects final answer generation to Qwen3.5-9B and retains the external model for memory construction and retrieval-related stages.
}
\label{tab:implementation_details}
\end{table*}

\subsection{Memory Construction and Retrieval}

Current memory maintains a 30-second rolling window with a 10-second core and at most 30 frames, providing a fast path for questions about the ongoing scene.
Short-term memory retains up to 1,000 recent events from the latest 30 minutes, including timestamps, representative frames, captions, and aligned speech.
Stable events are asynchronously consolidated into event-centric episodic and semantic stores, which constitute the long-term memory of LightMem-Ego.
We use Qwen3-Embedding-4B for textual representations and VLM2Vec-V2.0 for visual representations.

At query time, a deterministic router examines temporal, current/recent, speech, visual, entity, counting, summary, and follow-up cues.
It may activate one or multiple memory stores and falls back to an available store when the preferred source is not ready or current memory has become stale.
A retrieval planner then collects and fuses timestamped multimodal evidence before answer generation.
Retrieved timestamps, transcripts, event descriptions, and representative frames are packed into a compact evidence view for the answer model.

\subsection{Answer Generation Backends}

All language-model calls use an OpenAI-compatible client interface, allowing the answer backend to be switched without modifying the memory pipeline.
In the remote profile, the configured API model performs answer generation.
In the local profile, Qwen3.5-9B is served by vLLM with a 16,384-token context window, a maximum of 4,096 output tokens, and non-thinking mode enabled.
The local model generates the final response for current-, short-term-, and long-term-memory queries, whereas the external API remains responsible for memory construction, retrieval planning, event refinement, episodic generation, and multimodal captioning.
This hybrid profile is used when low time to first token (TTFT) is important: serving the answer model locally removes the network round trip and remote-service queueing from the generation path, while preserving the option to use stronger or larger API models for asynchronous memory operations.

\begin{table*}[t]
\centering
\footnotesize
\setlength{\tabcolsep}{5pt}
\renewcommand{\arraystretch}{1.08}
\begin{tabularx}{\textwidth}{@{}
>{\raggedright\arraybackslash}p{0.15\textwidth}
>{\raggedright\arraybackslash}X
>{\raggedright\arraybackslash}p{0.24\textwidth}
@{}}
\toprule
\textbf{Stage or object} & \textbf{Memory object or routing decision} & \textbf{Role in the trace} \\
\midrule
Live observation
& When the user arrives at the office, timestamped frames and audio enter \(\mathcal{M}_{cur}\). The rolling view captures the ongoing activity, such as checking messages, making coffee, or entering a meeting.
& Makes the latest scene queryable before event refinement or long-term indexing finishes. \\

Short-term event
& As the stream advances, each action is represented by one or more provisional micro-events in \(\mathcal{M}_{st}\). Captioning and transcript backfilling produce refined records with time spans, keyframes, actions, objects, and speech.
& Preserves concrete daily evidence and supports recent recall while the long-term memory is still being constructed. \\

Episodic consolidation
& Once a 30-second window is closed and all of its micro-events are refined, the system creates a long-term episode in \(\mathcal{M}_{lt}^{\mathrm{epi}}\). The episode retains the source event IDs and links each description to its timestamped visual and audio evidence.
& Converts transient event records into persistent, provenance-linked experience anchors. \\

Semantic consolidation
& Across repeated office-arrival episodes, \(\mathcal{M}_{lt}^{\mathrm{sem}}\) records the grounded routine \emph{arrive at office \(\rightarrow\) check messages \(\rightarrow\) make coffee \(\rightarrow\) join the morning meeting}, together with links to supporting episodes.
& Represents a recurring routine without losing the concrete observations from which it was inferred. \\

Query routing
& For \emph{What did I usually do after arriving at the office this week?}, the long-horizon intent prioritizes \(\mathcal{M}_{lt}\). The planner retrieves semantic routine evidence and the relevant episodic anchors using text--visual hybrid retrieval.
& Matches the query horizon to long-term semantic and episodic evidence instead of scanning the live buffer. \\

Readiness gate
& If long-term construction is still in progress, the router exposes that state and, when the query permits, falls back to refined \(\mathcal{M}_{st}\) events. If \(\mathcal{M}_{lt}\) is ready, duplicate records from multiple retrieval paths are merged by canonical segment ID.
& Keeps the demo answerable during asynchronous updates and prevents repeated evidence from dominating the prompt. \\

Grounded readout
& The packed evidence contains the routine fact plus its supporting office-arrival episodes and timestamps. The answer model returns: \textbf{You usually checked messages, made coffee, and joined the morning meeting.}
& Produces the final answer from a compact, traceable evidence view rather than the raw week-long stream. \\
\bottomrule
\end{tabularx}
\caption{Concrete memory lifecycle and query-time trace for the week-long office-routine scenario in the LightMem-Ego demonstration. The trace illustrates how the same experience remains provenance-linked as it progresses from live observations to short-term events and long-term episodic and semantic memory.}
\label{tab:memory_lifecycle_trace}
\end{table*}

\subsection{Asynchronous Runtime}

Ingestion, ASR, visual encoding, memory refinement, consolidation, and query processing run in separate backend workers.
Current- and short-term-memory updates remain on the online path, while expensive refinement, embedding, indexing, and semantic consolidation are scheduled in the background.
This design allows incoming streams and interactive queries to proceed while a newly completed segment is still being enriched for long-term retrieval.

\begin{table*}[t]
\centering
\footnotesize
\setlength{\tabcolsep}{4.5pt}
\renewcommand{\arraystretch}{1.10}
\begin{tabularx}{\textwidth}{@{}
>{\raggedright\arraybackslash}p{0.15\textwidth}
>{\raggedright\arraybackslash}p{0.30\textwidth}
>{\raggedright\arraybackslash}X
@{}}
\toprule
\textbf{Deployment layer} & \textbf{Prototype placement} & \textbf{Interface and responsibility} \\
\midrule
Client capture & Phone/browser camera and microphone, or Rokid Glass with an Android companion & Compresses and timestamps frames/audio; retains the assigned \texttt{session\_id}; performs no VLM or LLM inference \\
Public API & FastAPI/Uvicorn on the backend server & Creates sessions, accepts multipart media, reports status, enqueues queries, and streams answer events over SSE \\
Media transport & Direct HTTP frame--audio upload; optional SRS-based RTMP/WHIP & Normalizes different client transports into the same timestamped ingestion interface \\
Online workers & Separate stream, ASR, refinement, consolidation, visual, memory, and query processes on the same server & Consume persistent task queues and update session-local memory state asynchronously \\
Main runtime & Python environment for the API, indexing, retrieval, and memory workers & Hosts the general application stack and OpenAI-compatible clients \\
Shared GPU runtime & GPU-backed workers, WhisperX, embedding services, and local vLLM share one A100 40GB GPU in the current demo & Provides a compact single-GPU worker deployment for the current demonstration workload \\
Optional query scaling & Query worker and local multimodal answer service can be moved together to the second A100 40GB GPU & Supports more image-intensive queries or higher query concurrency without competing with streaming memory workers for GPU memory \\
Session storage & Session-scoped directories plus persistent task queues and retrieval indexes & Separates concurrent users/devices and preserves provenance across current, short-term, and long-term memory \\
Configuration & Environment variables and a runtime-selectable answer profile & Controls ports, model endpoints, GPU placement, retrieval settings, and local-versus-remote generation without changing clients \\
\bottomrule
\end{tabularx}
\caption{Deployment layout of the LightMem-Ego demonstration. The current prototype co-locates its worker stack on one A100 GPU; a second GPU is an optional scaling configuration for image-intensive query-time inference rather than a requirement of the default deployment.}
\label{tab:deployment_details}
\end{table*}

\section{Memory Lifecycle and Routing}

\subsection{From Live Observations to Long-Term Memory}

LightMem-Ego uses a deterministic, readiness-gated lifecycle rather than asking a language model to decide when every observation should be remembered.
For a timestamped observation \(x_t\), the main state transition is
\begin{equation}
\begin{aligned}
x_t
&\rightarrow \mathcal{M}_{cur}, \\
\mathcal{M}_{cur}
&\rightarrow e_i^{\mathrm{prov}} \in \mathcal{M}_{st}
\rightarrow e_i^{\mathrm{ref}}, \\
e_i^{\mathrm{ref}}
&\rightarrow z_i^{\mathrm{epi}} \in \mathcal{M}_{lt}^{\mathrm{epi}}, \\
\{z_i^{\mathrm{epi}}\}_{i\in\mathcal{I}}
&\rightarrow s_j^{\mathrm{sem}} \in \mathcal{M}_{lt}^{\mathrm{sem}}.
\end{aligned}
\end{equation}
Here, \(\mathcal{I}\) indexes the episodic anchors that support a semantic fact.
These arrows denote materialization and consolidation rather than destructive movement: the same source interval remains linked across memory levels through timestamps, event identifiers, and keyframes.

Incoming frames and partial transcripts first update the rolling current-memory view.
As time advances, they are grouped into provisional short-term micro-events that are immediately searchable, even before expensive captioning completes.
An asynchronous refinement worker then attaches aligned transcript segments and structured visual fields, changing each event from \textit{provisional} to \textit{refined}.
A closed 30-second window becomes eligible for episodic consolidation only when all overlapping micro-events are refined and none has been marked stale for reprocessing.
The resulting episode retains its source micro-event identifiers, time span, captions, transcript, entities, state changes, and representative keyframes, and is incrementally appended to long-term episodic memory.
Long-term semantic memory is subsequently derived from recurring, provenance-linked episodic evidence, representing routines, preferences, and stable relations without discarding their supporting event anchors.

The active views remain bounded: \(\mathcal{M}_{cur}\) rolls with the live stream, and \(\mathcal{M}_{st}\) keeps only its recent time-and-capacity window.
Events removed from the active short-term view remain in the short-term archive used by asynchronous consolidation, while consolidated episodes persist in \(\mathcal{M}_{lt}\).
If a late transcript changes an event, the affected window is marked dirty and can be refined and consolidated again with a new version.
The current prototype therefore provides bounded active views and versioned reconstruction, but not a learned forgetting or importance-based retention policy.

\subsection{Query-Aware Memory Routing}

Routing precedes retrieval and is separated into two levels.
First, the rule-based memory router classifies the query using temporal-scope and intent cues and activates the relevant stores.
Current-scene questions prioritize \(\mathcal{M}_{cur}\); recent-recall and counting questions can combine \(\mathcal{M}_{cur}\), \(\mathcal{M}_{st}\), and \(\mathcal{M}_{lt}\); and retrospective summaries or routine questions prioritize \(\mathcal{M}_{lt}\).
Speech, visual-attribute, temporal, and entity-tracking questions are routed according to their requested time span, while follow-up questions additionally use the interaction cache to recover referenced entities, time ranges, and memory identifiers.

Second, a retrieval planner chooses a text-only, visual-only, or hybrid policy for each enabled store and assigns store-specific candidate budgets.
The router is readiness-aware: stale current memory is disabled in favor of short-term memory, unavailable short-term memory falls back to long-term memory, and an unfinished long-term index can fall back to recent evidence when the query permits it.
Results from all enabled stores are normalized into a common evidence schema, deduplicated using canonical segment identifiers, and packed with timestamps and provenance before generation.
Thus, routing determines where to search, whereas evidence fusion determines which retrieved items reach the answer model.
Table~\ref{tab:memory_lifecycle_trace} gives a concrete end-to-end trace using the week-long office-routine scenario.

\section{Deployment Details}

\subsection{Server Deployment}

Our demonstration backend is deployed on an Ubuntu 22.04 server equipped with two NVIDIA A100 40GB GPUs and 256GB of system RAM.
The public service is implemented with FastAPI and Uvicorn, while ingestion, ASR, event refinement, consolidation, embedding, memory construction, and query answering run as separate worker processes.
Workers communicate through session-scoped files and persistent task queues, so an upload request only validates and records incoming data before scheduling expensive processing.
This organization isolates failures and allows expensive stages, particularly event refinement, to be replicated independently without changing the client API.

The deployment uses three Python environments to avoid incompatible model-serving dependencies.
The main environment runs the API and most memory and retrieval workers; a separate WhisperX environment contains the ASR-specific PyTorch and alignment stack; and an isolated vLLM environment serves the optional local Qwen answer model.
In the current demo configuration, all worker processes are co-located on the same server, and GPU-backed workers share a single A100.
This compact placement is sufficient for the current demonstration workload and leaves the second GPU optional rather than making it a deployment requirement.
For more image-intensive queries or higher query concurrency, the query worker and its local multimodal answer service can be placed on the second A100, since multimodal generation accounts for most of the query-time GPU memory growth.
GPU assignments remain environment-configurable and can also be redistributed across separate hosts.
Hosted multimodal and language models are accessed through OpenAI-compatible endpoints, so the same deployment can use either public APIs or privately hosted services.

\subsection{Client and Session Interface}

Phone and browser clients create a session through the general stream API and upload JPEG, PNG, or WebP frames together with short audio chunks using multipart HTTP requests.
Rokid Glass uses a dedicated API namespace: the glasses SDK or an Android companion converts camera output to compressed images, packages audio as WAV or supported PCM, and forwards both modalities to the backend.
The backend itself does not depend on the proprietary glasses SDK, and all input modes enter the same current-, short-term-, and long-term-memory pipeline after ingestion.
An optional RTMP/WHIP path can be enabled through an SRS media server, although the direct frame--audio connector is used for the wearable demonstration because it requires less deployment infrastructure.

Each stream receives a unique \texttt{session\_id}, which is retained by the client and attached to subsequent upload, status, and query requests.
Multi-session mode allows phone, browser, and glasses sessions to run concurrently over a shared worker pool without a newly started stream cancelling earlier tasks.
Answers are returned through Server-Sent Events (SSE), enabling the interface to display queue, retrieval, and token-streaming updates before the complete response is available.
Session metadata may additionally include owner and device identifiers, but these fields do not alter the memory and retrieval logic.

\subsection{Operation and Portability}

The API, workers, embedding services, and local answer server are started independently and expose health or runtime-state records for monitoring.
Worker launchers reuse healthy services, record process identifiers and logs, and support multiple refinement workers for higher ingestion throughput.
The answer profile can be changed between the remote endpoint and local Qwen without changing client requests or stored memories; only the affected workers need to be restarted.

Runtime artifacts are organized by session and include uploaded media, current and short-term state, task records, consolidated memories, and retrieval indexes.
Model weights, API credentials, runtime sessions, queues, and logs are excluded from the source release and supplied through local paths and environment variables.
The prototype is intended for a controlled demonstration network: production deployment additionally requires authentication, TLS termination, access control, encrypted storage, and explicit retention policies.
Table~\ref{tab:deployment_details} summarizes the resulting deployment layout.

\end{document}